\documentclass[journal]{IEEEtran}

\ifCLASSINFOpdf
\else
   \usepackage[dvips]{graphicx}
\fi
\usepackage{url}

\usepackage{graphicx}
\usepackage{amsmath}
\usepackage{amssymb}
\usepackage{booktabs}
\usepackage{multirow} 
\usepackage{comment}
\usepackage{xcolor}
\usepackage{multicol} 

\usepackage[colorlinks,
linkcolor=blue,
anchorcolor=black,
citecolor=blue
]{hyperref}

\begin{document}

\title{PVINet: Point-Voxel Interlaced Network for Point Cloud Compression}



\author{Xuan Deng, Xingtao Wang, Xiandong Meng,

Xiaopeng Fan,~\IEEEmembership{Senior Member, IEEE}, and Debin Zhao

\thanks{Xuan Deng, Xiaopeng Fan and Debin Zhao are with the School of Computer Science and Technology, Harbin Institute of Technology, Harbin 150001, China, and also with the Peng Cheng Laboratory, Shenzhen 519055, China (e-mail: dengx01@pcl.ac.cn; fxp@hit.edu.cn; dbzhao@hit.edu.cn).}
\thanks{Xingtao Wang is with the School of Computer Science and Technology, Harbin Institute of Technology, Harbin 150001, China (e-mail: xtwang@hit.edu.cn)}
\thanks{Xiandong Meng is with the Peng Cheng Laboratory, Shenzhen 519055, China (e-mail: Mengxd@pcl.ac.cn)}
}

\markboth{IEEE SIGNAL PROCESSING LETTERS, VOL. 31, 2025}
{Shell \MakeLowercase{\textit{et al.}}: Bare Demo of IEEEtran.cls for IEEE Journals}
\maketitle
\begin{abstract}

In point cloud compression, the quality of a reconstructed point cloud relies on both the global structure and the local context, with existing methods usually processing global and local information sequentially and lacking communication between these two types of information. In this paper, we propose a point-voxel interlaced network (PVINet), which captures global structural features and local contextual features in parallel and performs interactions at each scale to enhance feature perception efficiency. Specifically, PVINet contains a voxel-based encoder ($\mathcal{E}_v$) for extracting global structural features and a point-based encoder ($\mathcal{E}_p$) that models local contexts centered at each voxel. Particularly, a novel conditional sparse convolution is introduced, which applies point embeddings to dynamically customize kernels for voxel feature extraction, facilitating feature interactions from $\mathcal{E}_p$ to $\mathcal{E}_v$. During decoding, a voxel-based decoder employs conditional sparse convolutions to incorporate point embeddings as guidance to reconstruct the point cloud.
Experiments on benchmark datasets show that PVINet delivers competitive performance compared to state-of-the-art methods.
\end{abstract}


\IEEEpeerreviewmaketitle

\section{Introduction}

\IEEEPARstart{P}{oint} clouds are crucial for applications like autonomous driving, robotics, human modeling, and physics simulations. While LiDAR technology has made their capture easier, it also poses challenges in storage and transmission. Point cloud compression addresses these issues by reducing costs while preserving essential geometric information.


Traditional point cloud compression predominantly employs octree-based structures exemplified by G-PCC~\cite{schwarz2018emerging,schnabel2006octree} with handcrafted encoding rules, while V-PCC~\cite{schwarz2018emerging} adapts video coding paradigms through 3D-to-2D projection.

Recently, learning-based point cloud compression methods have achieved great success, including point-based methods~\cite{you2021patch,you2022ipdae,wiesmann2021deep,thomas2019kpconv,he2022density,gu20203d,li2024hierarchical}, voxel-based methods~\cite{quach2019learning,wang2021multiscale,wang2022sparse,wu2024geometric}, octree-based methods~\cite{sun2024enhancingB,de2018distance,souto2021set} and some hybrid methods~\cite{pang2022grasp,pang2024pivot,zhang2024deeppcc}, etc. 
Point-based methods have introduced several novel convolution operations, such as KPConv~\cite{thomas2019kpconv,wu2019pointconv}, to extract local features from point clouds. These convolution operations effectively preserve the local density of point clouds. 
Voxel-based approaches leverage sparse convolutional networks (e.g., PCGCv2~\cite{wang2021multiscale} and its enhanced variant SparsePCGC~\cite{wang2022sparse}) to model global structure through structured volumetric processing, achieving superior compression ratios while maintaining structural integrity.
Hybrid frameworks combine point-based and voxel-based representations, integrating self-attention mechanisms to fuse global and local features better. Additionally, octree-based methods are widely adopted for lossless point cloud compression, demonstrating strong performance in compressing sparse point cloud data. 

Point-based methods effectively capture local geometric details but struggle with large-scale point clouds. Voxel-based approaches efficiently preserve global structures through volumetric compression. To balance both strengths, hybrid frameworks combine these paradigms: GRASP-Net~\cite{pang2022grasp} and PIVOT-Net~\cite{pang2024pivot} cascade point-voxel operations, while DeepPCC~\cite{zhang2024deeppcc} integrates voxel convolutions with point-based attention mechanism. Nevertheless, these cascaded methods process global and local information sequentially, thus lacking communication between these two types of information.

In this paper, we propose a point-voxel interlaced network (PVINet), which captures global structural and local contextual features in parallel and performs interactions at each scale to enhance feature perception efficiency. Specifically, PVINet contains a voxel-based encoder ($\mathcal{E}_v$) and a point-based encoder ($\mathcal{E}_p$). $\mathcal{E}_v$ extracts global structure features of the point cloud by voxelizing the input point cloud, while $\mathcal{E}_p$ models local contexts centered on each voxel. During encoding,  $\mathcal{E}_p$ integrates representations extracted by $\mathcal{E}_v$ to facilitate point embeddings to achieve structure-aware modeling of the local details. Meanwhile, $\mathcal{E}_v$ incorporates the point embeddings learned by $\mathcal{E}_p$ through conditional sparse convolutions to capture the structures conditioned on local contexts. During decoding, a voxel-based decoder employs conditional sparse convolutions to incorporate point embeddings as guidance to reconstruct the point cloud. The main contributions can be summarized as follows:

$\bullet$ A point-voxel interlaced network (PVINet) is designed to capture global structural features and local contextual features in parallel and perform interactions at each scale.

$\bullet$ A conditional sparse convolution is introduced to facilitate the interactions between structural information in voxel embeddings and geometric contexts in point embeddings.

$\bullet$ Extensive experiments are conducted to demonstrate the superiority of PVINet over state-of-the-art methods, especially in preserving details and preventing excessive smoothing.

\section{Methodology}
\begin{figure*}[t]
\centering
\includegraphics[width=0.9\textwidth]{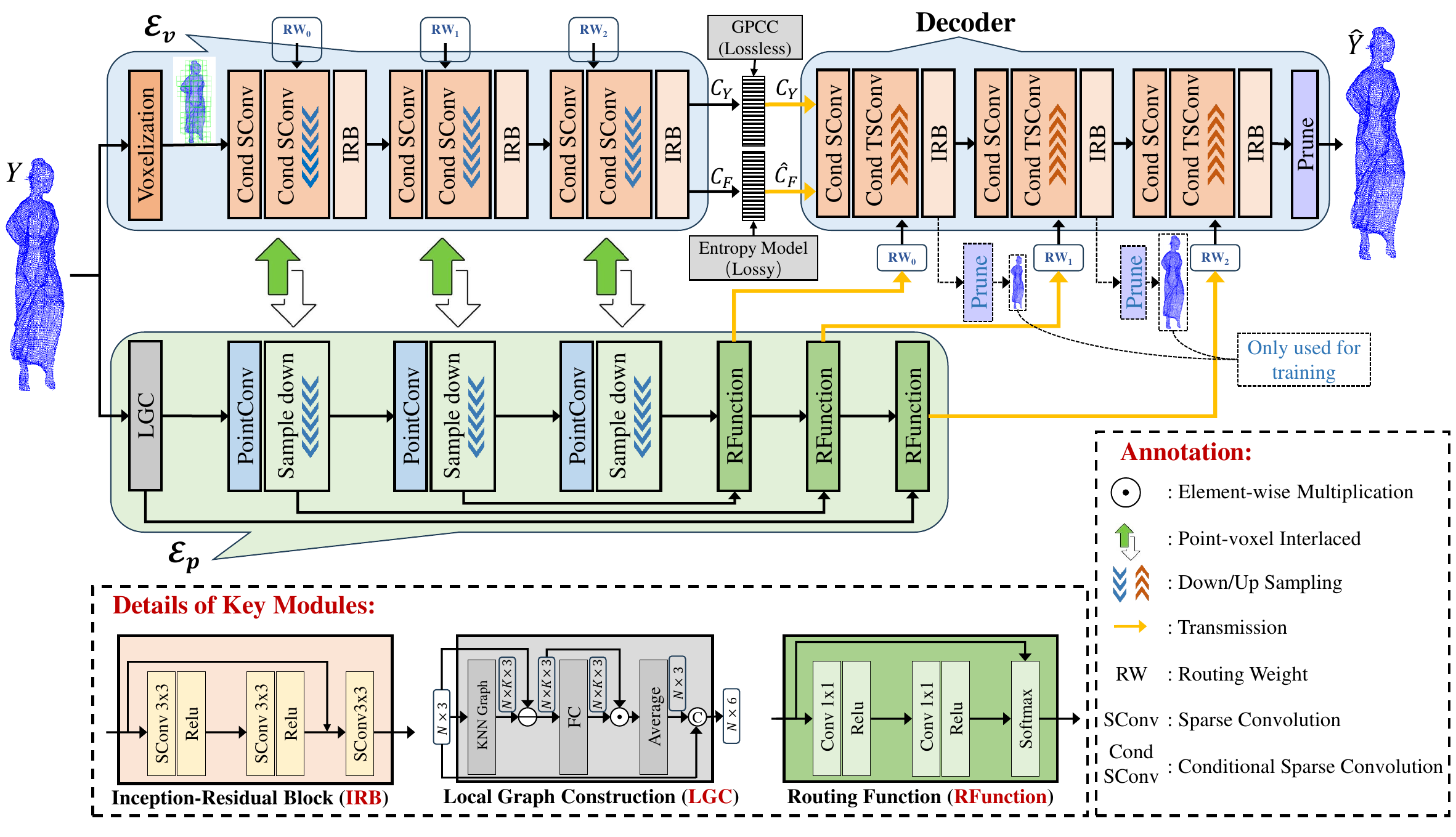}
\caption{Overview of the PVINet. The framework consists of three key modules: a point-voxel interlaced encoder with parallel voxel-based encoder and point-based encoder, a conditional generation module featuring three routing functions to determine routing weights and a decoder module.}
\label{fig: framework}
\end{figure*}

The proposed framework is illustrated in Fig.~\ref{fig: framework}. Initially, a point-voxel interlaced encoder processes the input point cloud to generate embeddings. Subsequently, entropy encoding is applied to facilitate arithmetic coding. Finally, a decoder is employed to reconstruct the point cloud with high fidelity.

\subsection{Point-Voxel interlaced Encoder}
The point-voxel interlaced encoder seamlessly integrates a point-based encoder ($\mathcal{E}_p$) and a voxel-based encoder ($\mathcal{E}_v$) operating in parallel, as illustrated in Fig~\ref{fig: framework}.

\textbf{Voxel-based Encoder:} After voxelizing the point cloud, $\mathcal{E}_v$ leverages well-studied 3D sparse convolutions to capture the global object structure. 
Specifically, $\mathcal{E}_v$ extracts voxel features at three scales. 
In each scale, two 3D conditional sparse convolutions and an inception-residual block (IRB, see Fig.~\ref{fig: framework}), are employed for feature extraction. 
Sparse 3D convolutions facilitate the enlargement of the receptive field, effectively covering the entire object to model its overall structure.

\textbf{Point-based Encoder:}
To compensate for the loss of local geometric contexts in the $\mathcal{E}_v$, $\mathcal{E}_p$ is developed to capture these complementary cues. $\mathcal{E}_p$ has three stages to match the voxel-based encoder. First, the centers of the voxels are considered as anchor points. Then, for each point $p_i$ in a point cloud $P={\{p_i \in R^3,i=1,..., N\}}$, its offsets to $K$ nearest neighbors ($p_j-p_i$) are calculated and aggregated via a linear layer. In this way, the local contexts can be encoded in the embedding at the center point, as illustrated in the Local Graph Construction (LGC) block, shown in Fig.~\ref{fig: framework}. Subsequently, geometry-aware point convolutions~\cite{Wang2024Unsupervised} are adopted in the resulting point embeddings to capture the local contexts, which further facilitates $\mathcal{E}_p$ to better capture the local contexts.

\textbf{Voxel$\xrightarrow{}$Point Feature Interaction:}
To propagate the global structure cues learned by the $\mathcal{E}_v$ to the $\mathcal{E}_p$, trilinear interpolation is employed to retrieve features from $F^{voxel}_{stage 1}\in\mathbb{R}^{H\times W\times L \times C}$ using the coordinates of the points ($\mathbb{R}^{N \times 3}$), which are then added to $F^{point}_{stage 1}\in\mathbb{R}^{N \times C}$. In this way, the structure information can be incorporated by $\mathcal{E}_p$ to better model local contexts. 

\begin{figure}[t]
    \centering
    \includegraphics[width=0.4\textwidth]{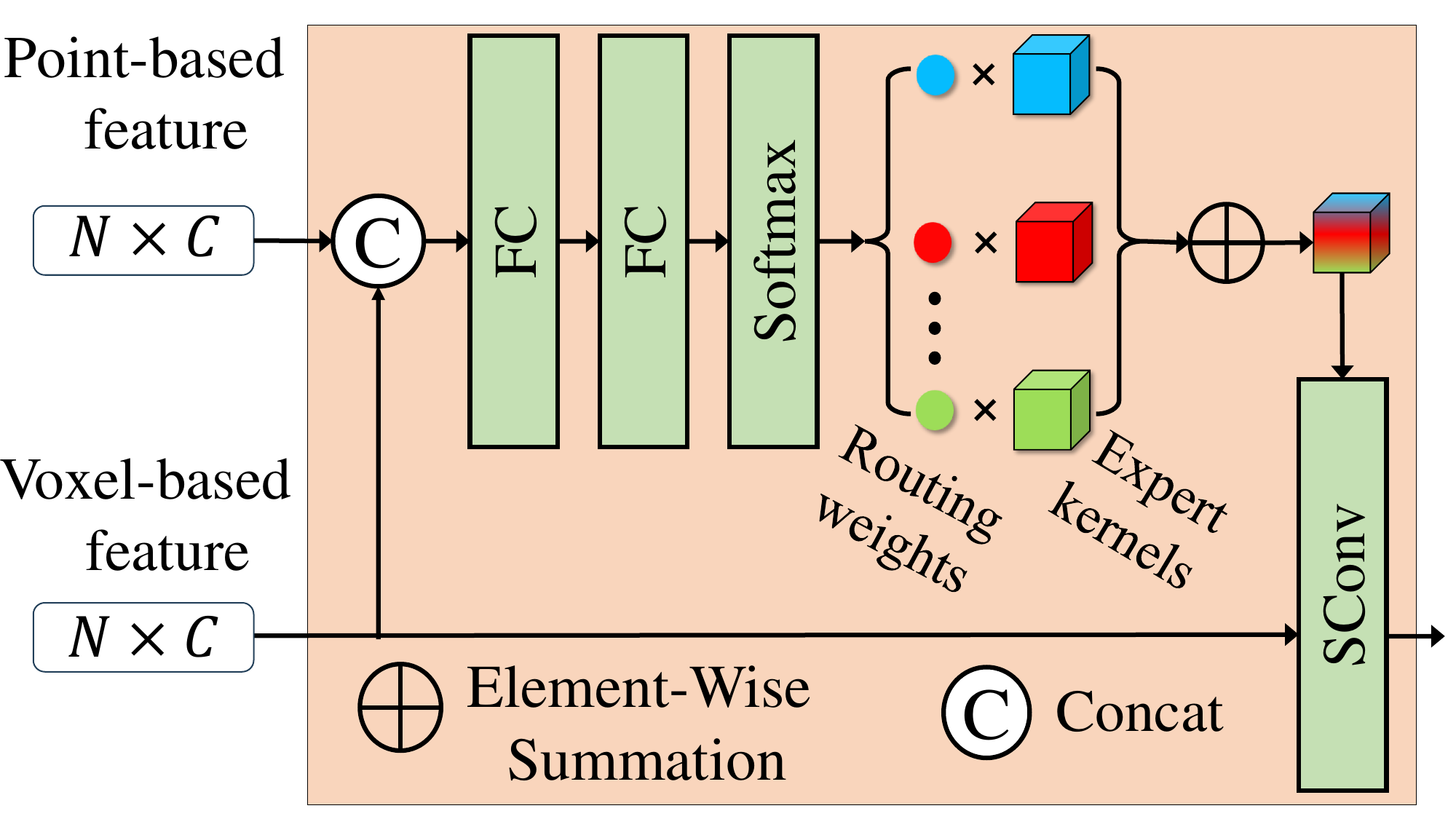}
    \caption{An illustration of the conditional sparse convolution, where different colors denote expert kernels, linearly combined based on routing weights derived from point-based and voxel-based features via Softmax.}
\label{fig:condsconv}
\end{figure}

\textbf{Point$\xrightarrow{}$Voxel Feature Interaction:}
To incorporate local contexts to better understand the global structure, $F^{point}_{stage 1}\in\mathbb{R}^{N \times C}$ are first concatenated with $F^{voxel}_{stage 1}\in\mathbb{R}^{H\times W\times L \times C}$ along the channel dimension and then conditional sparse convolutions are employed to customize kernels conditioned on $F^{point}_{stage 1}$, which is illustrated in Fig.~\ref{fig:condsconv}. Specifically, the point features are first pooled to obtain $Pool(F^{point}_{stage 1})\in\mathbb{R}^{1\times C}$. Then, as shown in the Routing Function (RFunction) module in Fig.~\ref{fig: framework}, an MLP and a softmax layer are used to predict a series of routing weights ${(\alpha_1,\alpha_2,...,\alpha_n)}$. Next, the resultant routing weights are used to assemble a kernel by combining a group of expert convolutional kernels ${(W_1,W_2,...,W_n)}$:
\begin{equation}
    W=\alpha_1W_1+\alpha_2W2+...+\alpha_nW_n.
\end{equation}
By parameterizing the convolutional kernels using point features, the local contexts can be incorporated to guide the $\mathcal{E}_v$ to better model the global structure. Unlike existing attention mechanisms~\cite{zhang2024deeppcc}, the conditional sparse convolution generates routing weights that vary with input point clouds, enabling dynamic adaptation to their characteristics.
\subsection{Arithmetic Encoding}
After the point-voxel interlaced encoder, the G-PCC codec ensures lossless encoding of voxel coordinates $C^{voxel}_{stage3}$, while entropy coding compresses voxel features $F_{stage3}^{voxel}$. During entropy encoding, $F_{stage3}^{voxel}$ is quantized and encoded using conditional entropy models~\cite{balle2018variational}, preserving global structural information. To convey local structures, a direct approach is encoding point coordinates $C^{point}_{stage3}$ and features $F^{point}_{stage3}$, but this incurs high memory and transmission costs. Instead, routing weights learned from point features are transmitted, providing complementary cues for decoding fine local details.

\begin{figure}[t]
    \centering
    \includegraphics[width=0.5\textwidth]{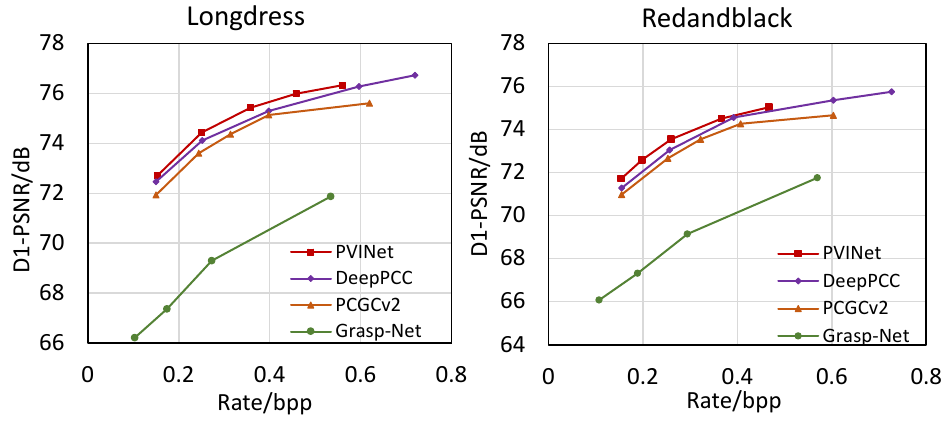}
    \caption{R-D performance curves of PVINet evaluated on the 8iVFB dataset.}
\label{fig: rd_curves}
\vspace{-10pt}
\end{figure}

\subsection{Decoder}
The decoder consists of three stages to progressively reconstruct the point cloud in a coarse-to-fine manner, as shown in Fig.~\ref{fig: framework}. At each stage, the voxel features are first passed to a conditional sparse convolution, a conditional transposed sparse convolution, and an IRB block under the guidance of the routing weights. In this way, the decoder can employ the local context information and facilitate it to reconstruct the point cloud better. It is worth noting that routing weights only introduce negligible coding bits.

\subsection{Loss Function}
The overall loss function includes distortion and rate losses:
\begin{equation} 
L=L_{Dist} + \lambda L_{Rate},
 \label{eq:loss_function}
\end{equation}
where $L_{Dist}$ penalizes the distortion of the decoded point cloud and $L_{Rate}$ penalizes the bit rate. Following~\cite{balle2018variational}, the quantization step is replaced with an additive uniform noise, and the number of bits is estimated as the rate loss. $\lambda$ is a parameter to control the trade-off between bit rate and distortion.

\textbf{Distortion Loss:} 
Binary cross entropy loss is employed as the distortion loss in the experiments:
\begin{equation} 
L_{Dist}=\frac{1}{N}\sum\nolimits_i -(x_i log(p_i)+(1-x_i)log(1-p_i)),
\label{eq:lossD_function}
\end{equation}
where $x_i$ is the voxel label (1 for occupied and 0 for empty) and ${p_i}$ is the occupancy probability of the voxel. 

\begin{table*}[ht]
    \centering
    \caption{BD-Rate gains measured using both D1 PSNR(\%) and D2 PSNR(\%) for PVINet (ours) against the G-PCC (octree), GRASP-Net~\cite{pang2022grasp}, HPSR-PCGC~\cite{li2024hierarchical},PCGCv2~\cite{wang2021multiscale},DeepPCC~\cite{zhang2024deeppcc}, SparsePCGC~\cite{wang2022sparse} for lossy coded dense point clouds. The larger the negative percentage, the greater the gain of our method.}
    \label{tab:mytable4}
    \small
    \resizebox{\linewidth}{!}{
    \renewcommand{\arraystretch}{0.75}  
     \begin{tabular}{c c c c c c c c c c c c c c}  
        \toprule
        {}&{}&\multicolumn{2}{c}{\textbf{Gain over}}&\multicolumn{2}{c}{\textbf{Gain over}}&\multicolumn{2}{c}{\textbf{Gain over }}&\multicolumn{2}{c}{\textbf{Gain over}}&\multicolumn{2}{c}{\textbf{Gain over }}&\multicolumn{2}{c}{\textbf{Gain over }}\\
        
        {}&{}&\multicolumn{2}{c}{\textbf{G-PCC(octree)}}&\multicolumn{2}{c}{\textbf{GRASP-Net~\cite{pang2022grasp}}}&\multicolumn{2}{c}{\textbf{HPSR-PCGC~\cite{li2024hierarchical}}}&\multicolumn{2}{c}{\textbf{PCGCv2~\cite{wang2021multiscale}}}&\multicolumn{2}{c}{\textbf{DeepPCC~\cite{zhang2024deeppcc}}}&\multicolumn{2}{c}{\textbf{SparsePCGC~\cite{wang2022sparse}}}\\
        \midrule
        \textbf{Dataset}&\textbf{Point Cloud} & \textbf{D1 (\%)} & \textbf{D2 (\%)} & \textbf{D1 (\%)} & \textbf{D2 (\%)} &\textbf{D1 (\%)} & \textbf{D2 (\%)} & \textbf{D1 (\%)} & \textbf{D2 (\%)}& \textbf{D1 (\%)}& \textbf{D2 (\%)}& \textbf{D1 (\%)}& \textbf{D2 (\%)} \\ 
        \midrule
        { }&Andrew &          -90.4 & -79.2& -72.0 &  -64.6 &-21.4&-18.5& -1.0 &-7.8 &-3.8& -7.0&25.5& 20.23    \\ 
        {MUVB}&David &        -71.9 & -80.3& -68.9 &  -65.1 &-25.3&-20.2& -10.0 &-9.3 &-3.2 & -12.6& 20.32& 19.36   \\ 
        {(9bit)}&Phil &        -89.3 & -79.5& -68.8 &  -64.9 &-24.02&-21.3& -7.4 &-8.7 &0.2&  3.3&18.36& 16.36   \\ 
        {}&Sarah &            -84.0 & -79.0& -74.4 &  -63.9 &-18.92&-13.2& -5.1 &-9.7& -9.9& -6.2&22.36 &  20.32   \\ 
        \midrule
        {}&Longdress &        -92.0 & -83.7& -71.3 & -62.3   &-29.61&-25.85& -20.4 & -15.8& -11.3 & -6.0& 26.48 & 20.85 \\ 
        {8iVFB}&Loot &        -92.4 & -83.5&  -73.0 & -62.1  &-18.91&-26.35& -21.7 & -20.7& -12.6 & -9.4& 26.81 & 16.64 \\ 
        {(10bit)}&Redandblack & -95.9 & -80.9&  -58.9 & -23.2 &-27.4&-45.8& -17.8 & -14.1& -12.9 & -4.2& 17.96 & 10.63 \\ 
        {}&Soldier &          -90.8 & -80.9&  -69.7 & -60.7  &-29.0&-25.3& -18.7 & -17.5& -7.0 & -3.5& 26.29 & 19.97 \\
        \midrule
        {}&Basketball player & -95.6 & -88.8&  -79.8 & -77.6 &-32.6 &-28.2& -21.0 & -20.5& -7.1 & -5.8& 22.12 & 14.47  \\ 
        {OWlii}&Dancer &      -97.6 & -91.3&  -74.9 & -70.3 &-20.8 &-20.5 & -19.9 & -12.8 & -8.5 & -6.7 & 24.51 & 14.54 \\ 
        {(11bit)}&Exercixe &  -98.3 & -91.2&  -72.1 & -20.3 &-18.6 &-21.1 & -20.7 & -11.8 & -9.7 & -8.1& 24.19 & 14.96 \\ 
        {}&Model &            -96.1 & -86.5& -72.1 & -70.4  &-25.6 &-21.0& -20.7  & -12.5& -10.2 & -7.5& 20.11 & 4.15  \\ 
        \midrule
        {}&Average&            \textbf{-91.2} & \textbf{-79.0}& \textbf{-71.0} &  \textbf{-63.9}&\textbf{-24.1} &  \textbf{-22.05}& \textbf{-15.5} & \textbf{-15.4}& \textbf{-8.0} & \textbf{-6.1}& \textbf{22.9} & \textbf{16.0} \\
        \bottomrule
    \end{tabular}
    }
\end{table*}

\section{Experiments}

In this section, the performance of our method is first compared with existing point cloud compression techniques. Subsequently, ablation studies are conducted to highlight the effectiveness of each module design.

\subsection{Experiment Details}
\textbf{Datasets:}
The experiments are conducted using the ShapeNet dataset~\cite{chang2015shapenet}, which consists of approximately 51,300 CAD surface models. Following the approach in PCGCv2~\cite{wang2021multiscale}, dense sampling is applied, and $(x, y, z)$ coordinates are quantized to 7-bit values. For evaluation, 12 point clouds are selected: four from 8i Voxelized Full Bodies (8iVFB)~\cite{d20178i}, four from the Owlii dynamic human mesh~\cite{xu2017owlii}, and four from Microsoft Voxelized Upper Bodies (MVUB)~\cite{loop2016microsoft}. These point clouds, covering diverse scales and structures, are widely used in the MPEG and JPEG Pleno Common Test Conditions (CTC)~\cite{schwarz2018common, pcc2019jpeg}.



\textbf{Training Details:}
Five models are trained, each corresponding to a different $\lambda$ value (1e-1, 5e-1, 1, 2, 3) in Eq~\eqref{eq:loss_function}. All models are trained for 25 epochs with a batch size of eight, using the Adam optimizer and an initial learning rate of $1 \times 10^{-5}$. Notably, the downsampled coordinates ($C^{voxel}_{stage3}$) are excluded from the training process and are encoded losslessly using the G-PCC~\cite{schwarz2018emerging} method.

\textbf{Evaluation Metrics:}
The quantitative evaluation of our approach is based on RD criteria, using Bjøntegaard delta rate (BD-Rate) and BD-PSNR. To assess the quality of point cloud reconstruction, point-to-point PSNR (D1 PSNR) and point-to-plane PSNR (D2 PSNR)~\cite{schwarz2018emerging} are utilized. Bits per point (Bpp) is used as the compression ratio metric.

\begin{figure}[t]
    \centering
    \includegraphics[width=0.5\textwidth]{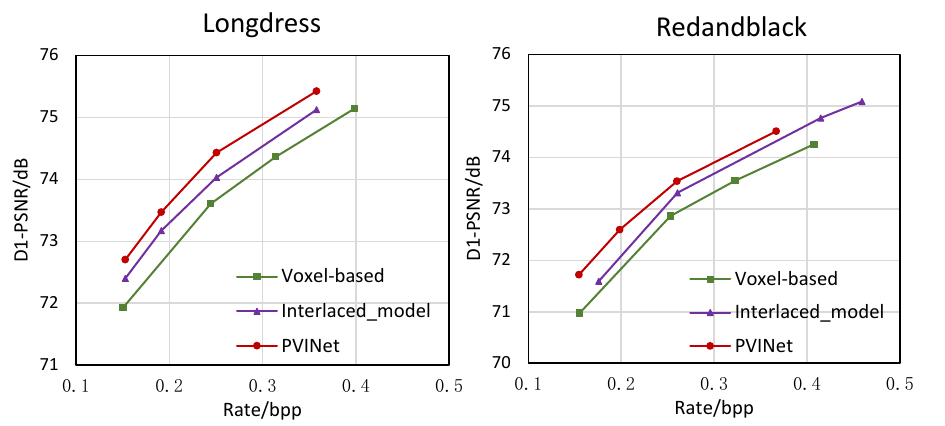}
    \caption{Ablation study on the Longdress and Redandblack datasets to evaluate the impact of the interlaced structure and Conditional Sparse Convolutions.}
\label{fig: Ablation study}
\end{figure}

\subsection{Experimental Results}
\textbf{Quantitative Results:}
Table~\ref{tab:mytable4} benchmarks the proposed method against conventional and learning-based compression approaches, including G-PCC~\cite{schwarz2018emerging}, GRASP-Net~\cite{pang2022grasp}, HPSR-PCGC~\cite{li2024hierarchical}, PCGCv2~\cite{wang2021multiscale}, DeepPCC~\cite{zhang2024deeppcc}, and SparsePCGC~\cite{wang2022sparse}. Evaluated through BD-Rate analysis with D1 (D2) PSNR and Bpp metrics, the PVINet achieves ${91.2\%}$ (${79.0\%}$) and ${71.0\%}$ (${63.9\%}$) bit rate reductions compared to G-PCC and GRASP-Net respectively, along with consistent BD-PSNR gains across dense point cloud datasets. Figure~\ref{fig: rd_curves} illustrates the rate-distortion performance through representative dataset samples. The interlaced network architecture combined with conditional sparse convolution enables ${15.5\%}$ and ${8\%}$ bit rate savings compared to PCGCv2 and DeepPCC respectively in D1 PSNR evaluation.
Compared to SparsePCGC, PVINet incurs a ${22.9\%}$ (${16.0\%}$) loss in D1 (D2) measurements. This is because, SparsePCGC employs a multistage strategy for lossless compression of downscaled thumbnail point clouds, while our method and other methods like PCGCv2 and DeepPCC directly use G-PCC for simplicity and compatibility.

\begin{figure}[t]
    \centering
    \includegraphics[width=0.45\textwidth]{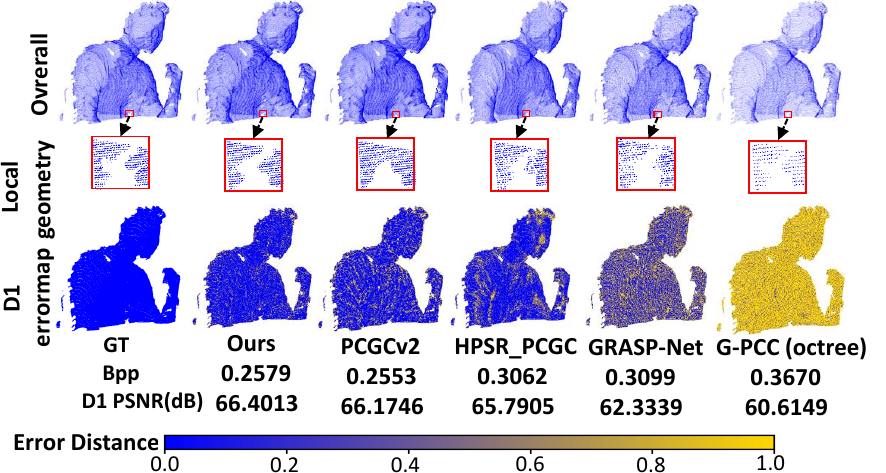}
    \caption{Visualization of geometric reconstruction results on David9. The final row exhibits error maps between the reconstructed point clouds and the ground truth in terms of D1 PSNR.}
\label{fig: subjective performance}
\vspace{-10pt}
\end{figure}

\textbf{Qualitative Results:}
Qualitative results in Fig.~\ref{fig: subjective performance} further confirm the efficacy of the proposed method, with reconstructed point clouds preserving both global structure and fine geometric details. In contrast, GRASP-Net retains rich details but exhibits high reconstruction errors due to structural inconsistencies, while PCGCv2 captures the overall skeleton well but suffers distorted details in the hand regions. Overall, the proposed approach strikes an optimal balance between structural fidelity and detail retention, achieving the lowest reconstruction error among all evaluated methods.



\subsection{Ablation Study}
This section presents an ablation study to evaluate the contributions of different components in the proposed framework. As illustrated in Fig.~\ref{fig: Ablation study}, the point-voxel interlaced structure is first evaluated by removing the point-based encoder $\mathcal{E}_p$ and comparing its performance with the interlaced model. The interlaced structure demonstrates a significant BD-Rate improvement of ${10.3\%}$ over the voxel-only method, highlighting its ability to combine the advantages of both voxel-based and point-based encodings for effective modeling of global structure and local details simultaneously.
Next, Conditional Sparse Convolutions are incorporated, and routing weight data is embedded into the bitstream to form the final PVINet model. This integration further enhances performance with a BD-Rate improvement of ${7.7\%}$, while maintaining efficient inference complexity. The results confirm that each individual module contributes to the overall performance improvement, underscoring the effectiveness of the proposed design.


\section{Conclusion}

This paper presents a novel point-voxel interlaced network (PVINet) for efficient point cloud compression. PVINet innovatively combines a voxel-based encoder and a parallel point-based encoder to jointly capture the global structural features and fine-grained local contexts of the input point cloud in an interlaced manner. The routing weights generated by the point-based encoder, along with the encoded feature stream, are transmitted to the decoder to reconstruct the overall structure while accurately recovering the detailed local geometry of the point cloud. Extensive experiments demonstrate that PVINet effectively preserves both the skeleton points and intricate local geometric details, achieving competitive performance on widely used benchmark datasets.

\clearpage

\twocolumn  
\bibliographystyle{IEEEtran}  
\bibliography{main}  

\vfill  

\end{document}